\documentclass[10pt,twocolumn,letterpaper]{article}

\usepackage{cvpr}
\usepackage{times}
\usepackage{epsfig}
\usepackage{graphicx}
\usepackage{amsmath}
\usepackage{amssymb}
\usepackage{booktabs}
\usepackage{mdwlist}

\def\ie{{\em i.e.}}
\def\eg{{\em e.g.}}
\def\etal{{\em et al.}}

\usepackage[breaklinks=true,bookmarks=false]{hyperref}

\cvprfinalcopy 

\def\cvprPaperID{****} 

\begin{document}

\title{Learning Semantic Neural Tree for Human Parsing}

\author{Ruyi Ji$^1$\thanks{Both authors contributed equally to this work.}, Dawei Du$^{2*}$, Libo Zhang$^{1}$\thanks{Corresponding author (libo@iscas.ac.cn). This work is supported by the National Natural Science Foundation of China under Grant No. 61807033, the Key Research Program of Frontier Sciences, CAS, Grant No. ZDBS-LY-JSC038, Youth Innovation Promotion Association CAS, and Outstanding Youth Scientist Project of ISCAS.}, \\
Longyin Wen$^3$, Yanjun Wu$^1$, Chen Zhao$^1$, Feiyue Huang$^4$, and Siwei Lyu$^2$\\
$^1$Institute of Software Chinese Academy of Sciences, China,\\
$^2$University at Albany, State University of New York, USA,\\
$^3$JD Finance America Corporation, USA, $^4$Tencent Youtu Lab, China. \\
{\tt\small \{ruyi2017,libo,yanjun,zhenchen\}@iscas.ac.cn, \{ddu,slyu\}@albany.edu}\\
{\tt\small  huangfeiyue@gmail.com, longyin.wen.cv@gmail.com}\\
}

\maketitle

\begin{abstract}
The majority of existing human parsing methods formulate the task as semantic segmentation, which regard each semantic category equally and fail to exploit the intrinsic physiological structure of human body, resulting in inaccurate results. In this paper, we design a novel semantic neural tree for human parsing, which uses a tree architecture to encode physiological structure of human body, and designs a coarse to fine process in a cascade manner to generate accurate results. Specifically, the semantic neural tree is designed to segment human regions into multiple semantic subregions (\eg, face, arms, and legs) in a hierarchical way using a new designed attention routing module. Meanwhile, we introduce the semantic aggregation module to combine multiple hierarchical features to exploit more context information for better performance. Our semantic neural tree can be trained in an end-to-end fashion by standard stochastic gradient descent (SGD) with back-propagation. Several experiments conducted on four challenging datasets for both single and multiple human parsing, \ie, LIP, PASCAL-Person-Part, CIHP and MHP-v2, demonstrate the effectiveness of the proposed method. Code can be found at \url{https://isrc.iscas.ac.cn/gitlab/research/sematree}.
\end{abstract}

\section{Introduction}
Human parsing aims to recognize each semantic part, \eg, arms, legs and clothes, which is one of the most fundamental and critical problems in analyzing human with various applications, such as video surveillance, human-computer interaction, and person re-identification.

With the development of convolutional neural networks (CNN) on semantic segmentation task, human parsing has obtained significant accuracy improvement recently. However, the majority of existing algorithms \cite{DBLP:journals/corr/ZhaoSQWJ16,DBLP:conf/eccv/GongLLCYL18,DBLP:conf/eccv/ChenZPSA18,DBLP:journals/corr/abs-1809-05996} formulate the task as semantic segmentation, \ie, assign each pixel with a class label, such as arm and leg, which regards each category equally and fails to exploit the intrinsic physiological structure of human body, leading to inaccurate results. For example, it is difficult to simultaneously distinguish the {\em torso}, {\em upper-arms}, {\em lower-arms}, and {\em background} pixels, especially in the cluttered scenarios.

\begin{figure}[t]
    \centering
    \includegraphics[width=0.95\linewidth]{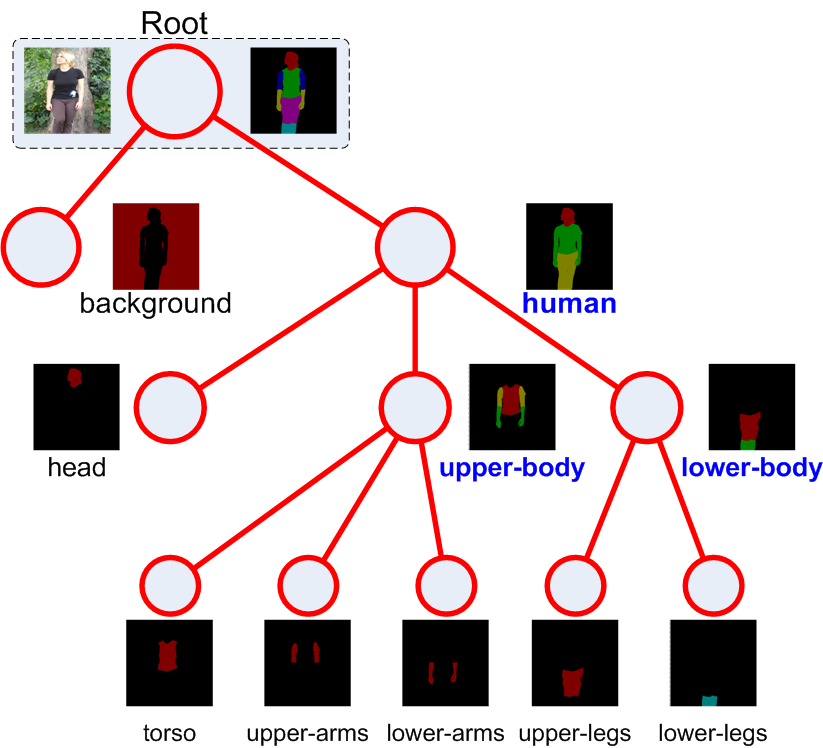}
    \caption{Category hierarchy used in the PASCAL-Person-Part dataset \cite{DBLP:conf/cvpr/ChenMLFUY14}. Seven labels are annotated in the dataset and the labels in blue font are defined as the virtual categories.}
    \label{fig:anno_tree}
\end{figure}

Inspired from human perception \cite{kimchi1992primacy}, we argue that the coarse to fine process in a hierarchical design is helpful to improve the performance of human parsing. As an example in Figure \ref{fig:anno_tree}, we introduce a virtual category {\em upper-body}, and first distinguish the {\em upper-body} from the {\em head} and {\em lower-body} pixels. After that, we segment the {\em torso}, {\em upper-arms}, and {\em lower-arms} regions from the segmented {\em upper-body} region. In this way, the hierarchical design in the cascade manner can generate more accurate results.

In this paper, we design a novel semantic neural tree (SNT) for human parsing, which uses a tree architecture to encode physiological structure of human body and design a coarse to fine process in a cascade manner. As shown in Figure \ref{fig:anno_tree}, it is natural to divide the virtual category \textit{human} into three parts including {\em head}, {\em upper-body}, and {\em lower-body}, each of which shares similar semantic context information. According to the topology structure of annotations in different datasets, we can design different tree architecture in a similar spirit. For the lead node of each path in the tree, our goal is to distinguish just a few categories. In general, the proposed semantic neural tree consists of four components, \ie, the backbone network for feature extraction, attention routing modules for sub-category partition, attention aggregation modules for discriminative feature representation and prediction modules for generating parsing result, laid in several levels. That is, we segment human regions into multiple semantic subregions in a hierarchical way using the attention routing module. After that, we introduce the semantic aggregation module to combine multiple hierarchical features to exploit rich context information. We generate the parsing result by aggregating the discriminative feature maps from each leaf node. Our SNT is trained in an end-to-end fashion using the standard stochastic gradient descent (SGD) with back-propagation \cite{DBLP:journals/neco/LeCunBDHHHJ89}.

Several experiments are conducted on four challenging datasets, \ie, LIP \cite{DBLP:journals/pami/LiangGSL19}, Pascal-Person-Part \cite{DBLP:conf/cvpr/ChenMLFUY14}, CIHP \cite{DBLP:conf/eccv/GongLLCYL18} and MHP-v2 \cite{DBLP:conf/mm/ZhaoLCSYF18}, demonstrating that our SNT method outperforms the state-of-the-art methods for both single and multiple human parsing. Meanwhile, we also carry out ablation experiments to validate the effectiveness of the components in our SNT. The main contributions are summarized as follows. (1) We propose a semantic neural tree for human parsing, which integrates the physiological structure of human body into a tree architecture, and design a coarse to fine process in a cascade manner to generate accurate results. (2) We introduce the semantic aggregation module to combine multiple hierarchical features to exploit rich context information.
(3) The experimental results on several challenging single and multiple human parsing datasets demonstrate that the proposed method surpasses the state-of-the-art methods.

\section{Related Work}
{\noindent \textbf{Semantic segmentation.}}
Semantic segmentation is one of the most relevant research directions to human parsing, which aims to assign each pixel with a class label, such as {\em car}, {\em flower}, and {\em person}. Several previous methods \cite{DBLP:journals/corr/ZhaoSQWJ16,DBLP:conf/cvpr/LinMSR17,DBLP:journals/pami/ChenPKMY18,DBLP:conf/cvpr/BilinskiP18,DBLP:conf/eccv/ChenZPSA18} use the fully convolutional network (FCN) to generate accurate segmentation results. Specifically, Zhao \etal~\cite{DBLP:journals/corr/ZhaoSQWJ16} propose the pyramid scene parsing network (PSPNet) to capture the capability of global context information by different-region based on context aggregation. In \cite{DBLP:conf/cvpr/LinMSR17}, the multi-path refinement network is developed to extract all the information available along the down-sampling process to enable high-resolution prediction using long-range residual connections. Besides, Chen \etal~\cite{DBLP:journals/pami/ChenPKMY18} introduce atrous spatial pyramid pooling (ASPP) to segment objects at multiple scales accurately. Improved from \cite{DBLP:journals/pami/ChenPKMY18}, they apply the depth-wise separable convolution to both ASPP and decoder modules to refine the segmentation results especially along object boundaries \cite{DBLP:conf/eccv/ChenZPSA18}. Recently, the meta-learning technique is applied in image prediction focused on the tasks of scene parsing, person-part segmentation, and semantic image segmentation, resulting in better performance \cite{DBLP:journals/corr/abs-1809-04184}. Bilinski and Prisacariu \cite{DBLP:conf/cvpr/BilinskiP18} propose a cascaded architecture with feature-level long-range skip connections, which incorporates the structure of ResNeXt's residual building blocks. However, these semantic segmentation methods are constructed without considering the relations among semantic sub-categories, resulting in limited performance for human parsing with fine-grained sub-categories.

{\noindent \textbf{Human parsing.}}
Furthermore, human parsing can be regarded as a fine-grained semantic segmentation task. To adapt to the human parsing task, more useful modules are proposed and combined in the semantic segmentation methods. Ruan~\etal~\cite{DBLP:journals/corr/abs-1809-05996} improve the PSPNet \cite{DBLP:journals/corr/ZhaoSQWJ16} by using the global context embedding module for multi-scale context information. Zhao~\etal~\cite{DBLP:conf/mm/ZhaoLCSYF18} employ three Generative Adversarial Network-like networks to perform semantic saliency prediction, instance-agnostic parsing and instance-aware clustering respectively. However, the aforementioned methods prefer to construct complex network for more discriminative representation, but consider little about semantic structure of human body when designing the network.

The semantic structure information is essential in human parsing. Gong \etal~\cite{DBLP:conf/eccv/GongLLCYL18} consider instance-aware edge detection to group semantic parts into distinct person instances. Liang~\etal~\cite{DBLP:journals/pami/LiangGSL19} propose a novel joint human parsing and pose estimation network, which imposes human pose structures into the parsing results without resorting to extra supervision. In~\cite{DBLP:conf/cvpr/Gong0LS0L19}, the hierarchical graph transfer learning is incorporated upon the parsing network to encode the underlying label semantic structures and propagate relevant semantic information. Different from them without exploring human hierarchy, we take full use of the category label hierarchy and propose a new tree architecture to learn semantic regions in a coarse to fine process.

{\noindent \textbf{Neural tree.}}
The decision tree (DT) is an effective model and widely applied in machine learning tasks. As the inherent of the interpretability, it is usually regard as an auxiliary tool to insight into the mechanism of neural network. However, the simplicity of identity function used in these methods means that input data is never transformed and thus each path from root to leaf node on the tree does not perform representation learning, limiting their performance. To integrate non-linear transformations into DTs, Kontschieder~\etal~\cite{7410529} propose the stochastic and differentiable decision tree model based neural decision forest. Similarly, Xiao \etal~\cite{DBLP:journals/corr/abs-1712-05934} develop a neural decision tree with a multilayer perceptron network at the root transformer. Different from above methods, our model pays more attention to the ``topology structure'' of annotations (see Figure \ref{fig:anno_tree}). That is, the proposed model have a flexible semantic topology depending on certain dataset. Moreover, we introduce the semantic aggregation module to combine multiple hierarchical features for more robustness.

\begin{figure*}[t]
    \centering
    \includegraphics[width=\linewidth]{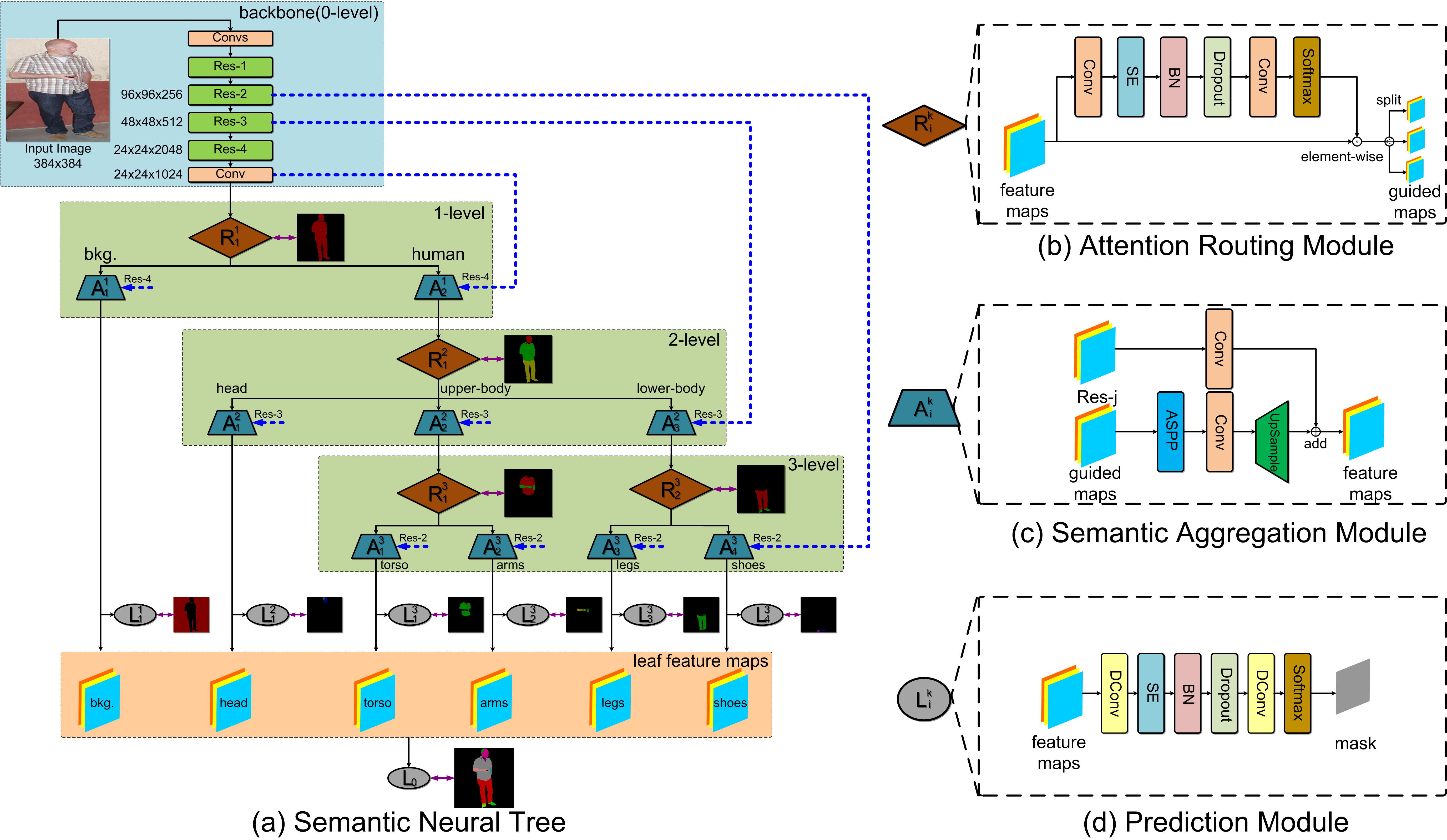}
    \caption{The tree architecture of our SNT model used on the LIP dataset \cite{DBLP:journals/pami/LiangGSL19}, which consists of four modules, \ie, the backbone network, (b) the attention routing module, (c) the semantic aggregation module, (d) the prediction module. The blue dashed lines indicate that the semantic aggregation modules in each level aggregate the features from different layers in the backbone. The purple double arrows denote the supervision for the attention routing and prediction modules. Best view in color.}
    \label{fig:structure}
\end{figure*}

\section{Methodology}
The goal of the proposed Semantic Neural Tree (SNT) method is to classify local parts of human along the path from root to leaf, and then fuse the feature maps before each leaf node to form the global representation for parsing prediction. We depart each sample $x\in{\it X}$ with the parsing label $y\in{\it Y}$. Notably, our model is not a full binary tree, because the topology of model is determined by the semantics of dataset. Based on our tree architecture, we group the parsing label into category label hierarchy. For example, as shown in Figure \ref{fig:structure}(a), the virtual category label \textit{head} consists of several child category labels \textit{face}, \textit{hair} and \textit{hat} in the LIP dataset \cite{DBLP:journals/pami/LiangGSL19}. Our model consists of four modules, the backbone network, the attention routing module, the semantic aggregation module, and the prediction module. We describe each module in detail in the following sections.

\subsection{Architecture}
{\noindent \textbf{Backbone network.}}
Similar to the previous works, we rely on residual blocks of ResNet-101 network \cite{DBLP:conf/cvpr/HeZRS16} to extract discriminative features of human in each sub-category. Our SNT can also work on other pre-trained networks, such as DenseNet \cite{DBLP:journals/corr/HuangLW16a} and Inception \cite{DBLP:journals/corr/SzegedyVISW15}.

Specifically, we remove the global average pooling and fully connected layers from the network and use the truncated ResNet-101 network \cite{DBLP:conf/cvpr/HeZRS16}, \ie, \text{Res}-$j, (j=1,2,3,4)$, as the backbone. Meanwhile, followed by the backbone, we add one convolutional layer with the kernel size $1\times1$ and stride size $1$ to reduce the channels of feature maps \text{Res}-$4$. Notably, as shown in Figure \ref{fig:structure}, we employ multi-scale feature representation as a powerful tool to improve the ResNet-101 backbone in the dense prediction task with highly localized discriminative regions in fine-grained categories.

{\noindent \textbf{Attention routing module.}}
After the backbone network, we need to solve how to split the tree structure. Given the sample $x$, in each level of the tree architecture, we employ the attention routing module to split the higher-level category labels and output the corresponding intermediate masks. That is, the $i$-th attention routing module at the $k$-th level $R_{i}^{k}$ is fed with the feature maps $\phi_{i}^{k-1}(x)$ at the $(k-1)$-th level. To this end, we supervise $R_{i}^{k}$ based on the labels of pre-set virtual categories.

As shown in Figure \ref{fig:structure}(b), the attention rounting module starts from one convolutional layer with the kernel size $1\times1$ and one Squeeze-and-Excitation (SE) layer \cite{DBLP:conf/cvpr/HuSS18}. Thus we can reduce the computational complexity and enforce the model to pay more attention to discriminative regions. After that, we use one dropout layer with the drop rate $0.5$, one convolutional layer with the kernel size $1\times1$ and one softmax layer to output the mask of the pixel-level human parts $\Psi_{i}^{k}(x)=\{\psi_{1}^{k}(x),\cdots,\psi_{I}^{k}(x)\}$ such that $\psi_{i}^{k}(x)\in[0,1]$. Notably, the channels of the mask consists of foreground channels and background channel, where $I$ denotes the channel number of $\Psi_{i}^{k}(x)$. The foreground channels denote the sub-category labels at node $i$ while background channel is defined as the other labels excluded from the sub-category labels at node $i$. With supervision on the masks, we can guide and split the feature maps at the $k$-th level into several semantic sub-categories, \ie, $\Phi_{i}^{k}(x)=\{\phi_{1}^{k}(x),\cdots,\phi_{I}^{k}(x)\}$.

{\noindent \textbf{Semantic aggregation module.}}
Followed by the attention routing module $R_{i}^{k}$, our goal is to extract discriminative feature representation for sub-categories. To this end, multi-scale feature representation is an important and effective strategy, \eg, skip-connections in the U-Net architecture \cite{DBLP:conf/eccv/ChenZPSA18}. On the other hand, the convolution with stride larger than one and the pooling operations will shrink feature maps, resulting in information loss in details such as the edge or small parts.

To alleviate these issues, we introduce the semantic aggregation module $A_i^k$ to deal with the feature maps $\phi_{i}^{k}(x)$. Specifically, we first adapt atrous spatial pyramid pooling (ASPP) \cite{DBLP:journals/pami/ChenPKMY18} to concatenate the features from multiple atrous convolutional layers with different dilation rates arranged in parallel. Specifically, the ASPP module is built to deal with the guided feature maps after the semantic router with dilation rates $[1, 6, 12, 18]$ to form multi-scale features. To aggregate multi-scale feature, we also use the upsampling layer to increase the spatial size of feature while halve the number of channels. After that, we use the addition operation to fuse the multi-scale features from the ASPP module and the residual features of the backbone \text{Res}-$j$ at the $j$-th stage (see Figure \ref{fig:structure}(c)). Thus we can learn more discriminative feature maps $\hat{\phi}_{i}^{k}(x)$ for prediction.

{\noindent \textbf{Prediction module.}}
Based on the feature maps after semantic aggregation $\hat{\phi}_{i}^{k}(x)$, we use the prediction modules $L_i^k$ in different levels to generate the parsing result for each sub-category. As shown in Figure \ref{fig:structure}(d), the prediction module includes one deformable convolutional layer \cite{DBLP:conf/cvpr/ZhuHLD19} with the kernel size $3\times3$, one SE layer \cite{DBLP:conf/cvpr/HuSS18}, one batch normalization layer, one dropout layer with drop rate $0.5$ and another deformable convolutional layer \cite{DBLP:conf/cvpr/ZhuHLD19} with the kernel size $3\times3$. Finally, the softmax layer is used to output an estimate for conditional distribution for each pixel. For each leaf node at the $k$-th level, we can predict the local part parsing result $\varphi_i^k(x)$.

Moreover, we combines all the feature maps of each leaf node $\hat{\phi}_{i}^{k}(x)$. Specifically, we remove the background channel in every leaf feature map and then concatenate the rest foreground channels, \ie, background, head, torso, arms, legs and shoes in Figure \ref{fig:structure}(a), such that the overall number of channels is equal to the number of categories. Thus we can predict the final parsing result $\mathcal{P}(x)$ by using the prediction module $L_0$.

\subsection{Loss function}
As discussed above, we use three loss terms on the attention routing module, each leaf node, and the final output after prediction modules to train the whole network in an end-to-end manner, which is computed as:
\begin{equation}
\begin{aligned}
\mathcal{L} &= \sum_{i}\sum_{k}\mathcal{L}_{R_i^k}(\Psi_i^k(x),\bar{y}_i^k) + \sum_{i}\sum_{k}\mathcal{L}_{L_i^k}(\varphi_i^k(x), \dot{y}_i^k) \\
&+ \mathcal{L}_{L_0}(\mathcal{P}(x), y^\ast),
\end{aligned}
\end{equation}
where $\mathcal{L}_{R_i^k}(\cdot,\cdot)$ denotes the cross-entropy loss between the masks $\Psi_i^k(x)$ generated by the attention routing module $R_i^k$ and the corresponding ground-truth $\bar{y}_i^k$ at the $k$-th level. $\mathcal{L}_{L_i^k}(\cdot,\cdot)$ denotes the cross-entropy loss between the output map $\varphi_i^k(x)$ by the leaf node and the corresponding ground-truth map $\dot{y}_i^k$ at the $k$-th level. $\mathcal{L}_{L_0}(\cdot,\cdot)$ denotes the cross-entropy loss between the final parsing result $\mathcal{P}(x)$ and the global parsing label $y^\ast$. It is worth noting that the channel number of $\bar{y}_i^k$ is equal to the number of sub-category labels of node $i$ at the $k$-th level, and the channel number of $y^\ast$ is equal to the total number of labels.

\subsection{Handling multiple human parsing}
To handle multiple human parsing, we integrate our method with the off-the-shelf instance segmentation framework, as similar as in \cite{DBLP:journals/corr/abs-1809-05996}. Specifically, we first employ the Mask R-CNN \cite{DBLP:conf/iccv/HeGDG17} pre-trained on MS-COCO dataset \cite{DBLP:conf/eccv/LinMBHPRDZ14} to segment human instances from images. Then, we train three SNT sub-models to obtain global and local human parsing results with different size of input images, \ie, one global sub-model and two local sub-models. Specifically, the global sub-model is trained on the whole images without distinguishing each instance; while the other two local sub-models are input by segmented instance patches from Mask R-CNN \cite{DBLP:conf/iccv/HeGDG17} and ground-truth respectively. Notably, we use the same architecture for the three sub-models. Finally, both the global and local results from these sub-models are combined to output multiple human parsing results by late fusion. That is, we concatenate the feature maps before leaf node on each sub-branches in our network. Followed by the prediction module, we can estimate the categories for each pixel under the supervision of cross-entropy loss function.

\section{Experiment}
Following the previous works \cite{DBLP:journals/corr/abs-1809-05996,DBLP:conf/eccv/GongLLCYL18,DBLP:journals/corr/abs-1804-03287,DBLP:conf/cvpr/Gong0LS0L19}, we compare our method with other state-of-the-arts on the validation set of two single human parsing datasets (\ie, LIP \cite{DBLP:journals/pami/LiangGSL19} and Pascal-Person-Part \cite{DBLP:conf/cvpr/ChenMLFUY14}) and two multiple human parsing datasets (\ie, CIHP \cite{DBLP:conf/eccv/GongLLCYL18} and MHP-v2 \cite{DBLP:conf/mm/ZhaoLCSYF18}). First of all, we introduce the implementation details of our method and the evaluation metrics as follows. Then, we conduct the ablation study to demonstrate the effectiveness of the proposed modules in the tree architecture.

\subsection{Implementation Details}
We implement the proposed framework in PyTorch. The source code of the proposed method will be made publicly available after the paper is accepted. All models are trained on a workstation with a 3.26 GHz Intel processor, 32 GB memory, and one Nvidia V100 GPU.

Following the previous works, we adopt the ResNet-101 \cite{DBLP:conf/cvpr/HeZRS16} that is pre-trained on the ImageNet dataset \cite{5206848} as the backbone network. For a fair comparison, we set input size of images $384\times384$ for single person parsing while $473\times473$ for multiple person parsing. For data argumentation, we adopt the strategy of random scaling (from $0.5$ to $1.5$), random rotation, random croping and left-right flipping the training data. We use the SGD algorithm to train the network with $0.9$ momentum, and $0.00005$ weight decay. The learning rate is initialized to $0.001$ and declined by $0.5$ in every $30$ epochs. Notably, the warming up policy is applied for training. That is, we use the learning rate of $0.0001$ to warm up the model in the first $10$ epochs, and then increase learning rate up to $0.001$ linearly. The model is optimized in $200$ epochs, where the dropout operation is valid only in the training phase. The topology of the proposed network is designed based on the sub-categories in different datasets, which is described in the appendix in detail.

\subsection{Metrics}
First, we employ the mean IoU metric (\text{mIOU}) to evaluate the global-level predictions in single human parsing datasets (\ie, LIP \cite{DBLP:journals/pami/LiangGSL19} and Pascal-Person Part \cite{DBLP:conf/cvpr/ChenMLFUY14}). Then, we use three metrics (\ie, $\text{AP}^r$, $\text{AP}^p$ and $\text{PCP}$) to evaluate the instance-level predictions in multiple human parsing. The $\text{AP}^r$ score denotes the area under the precision-recall curve based on the limitation of different IoU thresholds (\eg, $0.5$, $0.6$, $0.7$) \cite{DBLP:conf/eccv/HariharanAGM14}. $\text{PCP}$ elaborates how many body parts are correctly predicted of a certain person \cite{DBLP:journals/corr/LiZWLLF17}. $\text{AP}^p$ computes the pixel-level IoU of semantic part categories within a person. Similar to the previous works, we use the metrics of \text{mIoU} and $\text{AP}^r$ to evaluate the performance on the CIHP dataset \cite{DBLP:conf/eccv/GongLLCYL18} while $\text{PCP}$ and $\text{AP}^p$ to evaluate the performance on the MHP-v2 dataset \cite{DBLP:conf/mm/ZhaoLCSYF18}.

\subsection{Single Human Parsing}
\begin{figure*}[t]
    \centering
    \includegraphics[width=0.95\linewidth]{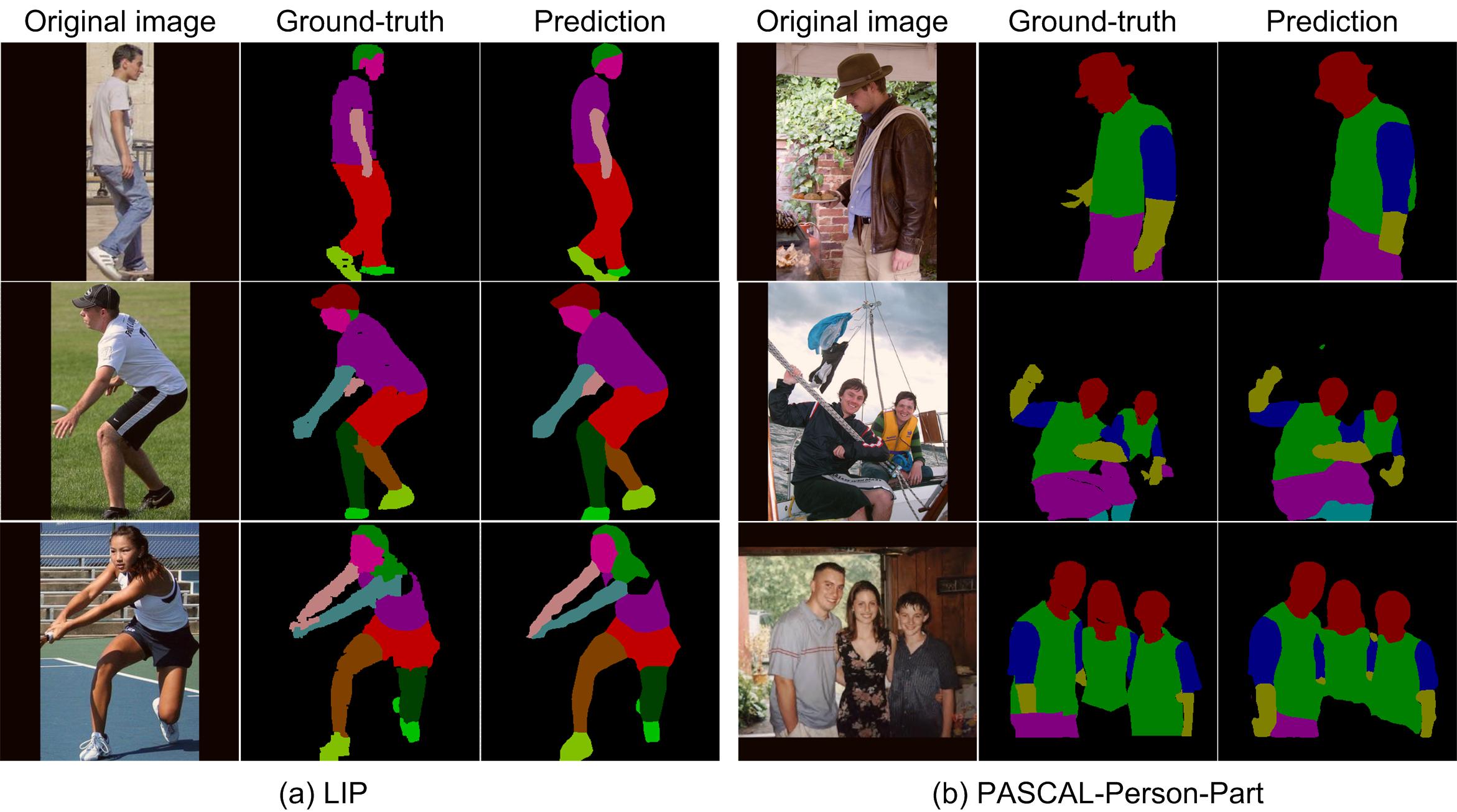}
    \caption{Some visualized examples for single human parsing: (a) the LIP dataset \cite{DBLP:journals/pami/LiangGSL19} and (b) the Pascal-Person-Part dataset \cite{DBLP:conf/cvpr/ChenMLFUY14}.}
    \label{fig:SHP_demo}
\end{figure*}

We compare the performance of single human parsing of our proposed method with other state-of-the-arts on the LIP \cite{DBLP:journals/pami/LiangGSL19} and Pascal-Person-Part \cite{DBLP:conf/cvpr/ChenMLFUY14} datasets. The qualitative human parsing results are visualized in Figure \ref{fig:SHP_demo}.

{\flushleft \textbf{Evaluation on LIP Dataset.}}
The LIP dataset defines $6$ body parts and $13$ clothes categories, including $50,462$ images with pixel-level annotations. Specifically, there are $19,081$ full-body images, $13,672$ upper-body images, $403$ lower-body images, $3,386$ head-missing images, $2,778$ backview images and $21,028$ images with occlusions. $30,462$ training and $10,000$ validation images are provided with publicly available annotations. As shown in Figure \ref{fig:structure}, we construct the tree architecture in $3$-level.

As presented in Table \ref{tab:LIP-1}, we can conclude that our method achieves the best performance in terms of all the three metrics. Since semantic segmentation methods (\eg, DeepLab \cite{DBLP:journals/pami/ChenPKMY18} and PSPNet \cite{DBLP:journals/corr/ZhaoSQWJ16}) consider little about fine-grained classification in the human parsing task, they perform not well. Moreover, the CE2P method \cite{DBLP:journals/corr/abs-1809-05996} improves PSPNet \cite{DBLP:journals/corr/ZhaoSQWJ16} by adding the context embedding branch, achieving $53.10$ \text{mIOU} score. Our method exceeds the current state-of-the-art CE2P \cite{DBLP:journals/corr/abs-1809-05996} by $1.63\%$ in terms of \text{mIOU} score. It indicates that our method can learn discriminative representation of each sub-category for human parsing. Moreover, as shown in Table \ref{tab:LIP-2}, our method obtain the best mIOU score in each sub-category. Notably, our method achieves considerable accuracy improvement compared with the other methods in some ambiguous sub-categories, \eg, \textit{glove}, \textit{j-suit}, and \textit{shoe}.

\begin{table}[t]
	\begin{center}
	\setlength{\tabcolsep}{5.5pt}	
	\caption{The evaluation results on the validation set of LIP \cite{DBLP:journals/pami/LiangGSL19}.}
	\vspace{2mm}
	\label{tab:LIP-1}
	\small{			
		\begin{tabular}{cccc}
			\toprule
			Method       						  		 	& pixel acc.  & mean acc.  & \text{mIoU}     \\
			\midrule
			Attention+SSL \cite{DBLP:conf/cvpr/GongLZSL17}  & -           & -          & 44.73    \\
			DeepLab \cite{DBLP:journals/pami/ChenPKMY18}       		 	& 84.09       & 55.62      & 44.80    \\
			MMAN \cite{Luo_2018_ECCV}          				& -           & -          & 46.81    \\
			SS-NAN \cite{DBLP:conf/iccv/ZhangTZLY17}        & 87.60       & 56.00      & 47.92    \\
			MuLA \cite{DBLP:conf/eccv/NieFY18}              & \textbf{88.50}       & 60.50      & 49.30    \\
			PSPNet \cite{DBLP:journals/corr/ZhaoSQWJ16}     & 86.23       & 61.33      & 50.56    \\
			JPPNet \cite{DBLP:journals/corr/abs-1804-01984} & 86.39       & 62.32      & 51.37    \\
			CE2P \cite{DBLP:journals/corr/abs-1809-05996}   & -           & -          & 52.56    \\
			CE2P(w flip) \cite{DBLP:journals/corr/abs-1809-05996}& 87.37       & 63.20      & 53.10    \\
			\midrule
			Ours   								& 88.05       & \textbf{66.42}      & \textbf{54.73}    \\
			\bottomrule
	    \end{tabular}
	}
	\end{center}
\end{table}

\begin{table*}
	\begin{center}
	\setlength{\tabcolsep}{1.5pt}
		\caption{The evaluation results on the validation set of LIP \cite{DBLP:journals/pami/LiangGSL19} in each category.}
		\vspace{2mm}
		\label{tab:LIP-2}
		\footnotesize{
		\begin{tabular}{ccccccccccccccccccccccccc}
			\toprule
		Method     & bkg.         & hat    & hair   & glove  & glasses & u-clothes & dress & coat   & socks   & pants   & j-suit   & scarf   & skirt   & face   & l-arm   & r-arm   & l-leg   & r-leg   & l-shoe   & r-shoe  & \text{mIoU}    \\
			\midrule
		Attention+SSL \cite{DBLP:conf/cvpr/GongLZSL17} &84.6 &59.8 &67.3 &29.0 &21.6 &65.3 &29.5 &51.9 &38.5 &68.0 &24.5 &14.9 &24.3 &71.0 &52.6 &55.8 &40.2 &38.8 &28.1 &29.0  &44.7 \\
		DeepLab~\cite{DBLP:journals/pami/ChenPKMY18}    & 84.1        & 59.8   & 66.2   & 28.8   & 23.9    & 65.0      & 33.7  & 52.9   & 37.7    & 68.0    & 26.1     & 17.4    & 25.2    & 70.0   & 50.4    & 53.9    & 39.4    & 38.3    & 27.0     & 28.4    & 44.8    \\
		PSPNet~\cite{DBLP:journals/corr/ZhaoSQWJ16}     & 86.1        & 63.5   & 68.0   & 39.1   & 23.8    & 68.1      & 31.7  & 56.2   & 44.5    & 72.7    & 28.7     & 15.7    & 25.7    & 70.8   & 59.7    & 62.3    & 54.9    & 54.5    & 42.3     & 42.9    & 50.6    \\
		MMAN~\cite{Luo_2018_ECCV}       & 84.8        & 57.7   & 65.6   & 30.1   & 20.0    & 64.2      & 28.4  & 52.0   & 41.5    & 71.0    & 23.6     & 9.7     & 23.2    & 69.5   & 55.3    & 58.1    & 51.9    & 52.2    & 38.6     & 39.0    & 46.8    \\
		JPPNet~\cite{DBLP:journals/corr/abs-1804-01984}     & 86.3        & 63.6   & 70.2   & 36.2   & 23.5    & 68.2      & 31.4  & 55.7   & 44.6    & 72.2    & 28.4     & 18.8    & 25.1    & 73.4   & 62.0    & 63.9    & 58.2    & 58.0    & 44.0     & 44.1    & 51.4    \\
		CE2P~\cite{DBLP:journals/corr/abs-1809-05996}       & 87.4        & 64.6   & 72.1   & 38.4   & 32.2    & 68.9      & 32.2  & 55.6   & 48.8    & 73.5    & 27.2     & 13.8    & 22.7    & 74.9   & 64.0    & 65.9    & 59.7    & 58.0    & 45.7     & 45.6    & 52.6    \\
			\midrule
		Ours       & \textbf{88.2}        & \textbf{66.9}   & \textbf{72.2}   & \textbf{42.7}   & \textbf{32.3}    & \textbf{70.1}      & \textbf{33.8}  & \textbf{57.5}   & 48.9   & \textbf{75.2}    & \textbf{32.5}     & \textbf{19.4}    & \textbf{27.4}    & \textbf{74.9}   & \textbf{65.8}    & \textbf{68.1}    & \textbf{60.3}    & \textbf{59.8}    & \textbf{47.6}     & \textbf{48.1}    & \textbf{54.7}    \\
			\bottomrule
	\end{tabular}}
	\end{center}
\end{table*}

{\flushleft \textbf{Evaluation on Pascal-Person-Part Dataset.}}
The PASCAL-Person-Part dataset \cite{DBLP:conf/cvpr/ChenMLFUY14} is originally from the PASCAL VOC-2010 dataset \cite{DBLP:journals/ijcv/EveringhamGWWZ10}, and then extended for human parsing with $6$ coarse body part labels (\ie, \textit{head}, \textit{torso}, \textit{upper-/lower-arms}, and \textit{upper-/lower-legs}). It consists of $1,716$ training images and $1,817$ testing images ($3,533$ images in total). As shown in Figure \ref{fig:anno_tree}, we construct the tree architecture in $3$-level. Specifically, the virtual category \textit{human} consists of three sub-categories, \ie, \textit{head}, \textit{upper-body} including torso, upper-arms and lower-arms and \textit{lower-body} including upper-legs and lower-legs.

We report the performance on the Pascal-Person-Part dataset in Table~\ref{tab:Pascal-Person-Part}. Similar to the trend in the LIP dataset \cite{DBLP:journals/pami/LiangGSL19}, the semantic segmentation methods, \eg, DeepLab \cite{DBLP:journals/pami/ChenPKMY18} and DeepLab v3+ \cite{DBLP:conf/eccv/ChenZPSA18}, perform inferior \text{mIoU} score, \ie, less than $68.00$. Moreover, the Graphonomy method \cite{DBLP:conf/cvpr/Gong0LS0L19} learns and propagates compact high-level graph representation among the labels within one dataset, resulting in better $69.12$ \text{mIoU} score. Besides, DPC \cite{DBLP:journals/corr/abs-1809-04184} achieves state-of-the-art performance with $71.34$ mIoU score. This is because it employs meta-learning to search optimal efficient multi-scale network for human parsing. Our SNT method obtains the best overall \text{mIoU} score of $71.59$ and best \text{mIoU} scores in terms of \textit{u-arms}, \textit{u-legs} and \textit{l-legs} among all the compared methods, which indicates the effectiveness of our proposed tree network.

\begin{table}[t]
	\begin{center}
	\setlength{\tabcolsep}{1.0pt}	
	\caption{The evaluation results on the validation set of Pascal-Person-Part \cite{DBLP:conf/cvpr/ChenMLFUY14}.}
	\vspace{2mm}
	\label{tab:Pascal-Person-Part}	
		\footnotesize{	
		\begin{tabular}{c|cccccccc}
			\toprule
			Method    										    & head        & torso  & u-arms & l-arms & u-legs & l-legs  & bkg.  & \text{mIoU}     \\
			\midrule
			HAZN~\cite{DBLP:conf/eccv/XiaWCY16}                 & 80.79       & 80.76  & 45.65  & 43.11  & 41.21  & 37.74   & 93.78       & 57.54    \\
			Attention+SSL \cite{DBLP:conf/cvpr/GongLZSL17}	    & 83.26       & 62.40  & 47.80  & 45.58  & 42.32  & 39.48   & 94.68       & 59.36    \\	
			Graph LSTM~\cite{DBLP:conf/eccv/LiangSFLY16}        & 82.69       & 62.68  & 46.88  & 47.71  & 45.66  & 40.93   & 94.59       & 60.16    \\
			SE LSTM~\cite{DBLP:journals/corr/LiangLSFYX17}      & 82.89       & 67.15  & 51.42  & 48.72  & 51.72  & 45.91   & 97.18       & 63.57    \\
			Part FCN~\cite{DBLP:conf/cvpr/XiaWCY17}             & 85.50       & 67.87  & 54.72  & 54.30  & 48.25  & 44.76   & 95.32       & 64.39    \\
            DeepLab \cite{DBLP:journals/pami/ChenPKMY18} 		&- &- &- &- &- &- &- &64.94\\
            MuLA \cite{DBLP:conf/eccv/NieFY18} 			        &- &- &- &- &- &- &- &65.10\\
            SAN \cite{huangsemantic}                            &86.12 &73.49 &59.20 &56.20 &51.39 &49.58 &96.01 &64.72\\
			WSHP~\cite{DBLP:journals/corr/abs-1805-04310}       & 87.15 & 72.28  & 57.07  & 56.21  & 52.43  & 50.36   & \bf{97.72}       & 67.60 \\
            DeepLab v3+ \cite{DBLP:conf/eccv/ChenZPSA18}        &- &- &- &- &- &- &- &67.84\\
            PGN \cite{DBLP:conf/eccv/GongLLCYL18}				&\bf{90.89} &\bf{75.12} &55.83 &\bf{64.61} &55.42 &41.57 &95.33 &68.40\\	
            Bilinski \etal~\cite{DBLP:conf/cvpr/BilinskiP18}	&- &- &- &- &- &- &- &68.60\\	
            Graphonomy \cite{DBLP:conf/cvpr/Gong0LS0L19} 		&- &- &- &- &- &- &- & 69.12\\	
			DPC~\cite{DBLP:journals/corr/abs-1809-04184}        & 88.81       & 74.54  & 63.85  & 63.73  & 57.24  & 54.55   & 96.66       & 71.34    \\
			\midrule
			Ours      										    & 89.15       & 74.76  & \bf{63.90}  & 63.95  & \bf{57.53}  & \bf{54.62}   & 96.84       & \bf{71.59}          \\
			\bottomrule
	     \end{tabular}
	     }
	\end{center}
\end{table}

\subsection{Multiple Human Parsing}
\begin{figure*}[t]
    \centering
    \includegraphics[width=0.95\linewidth]{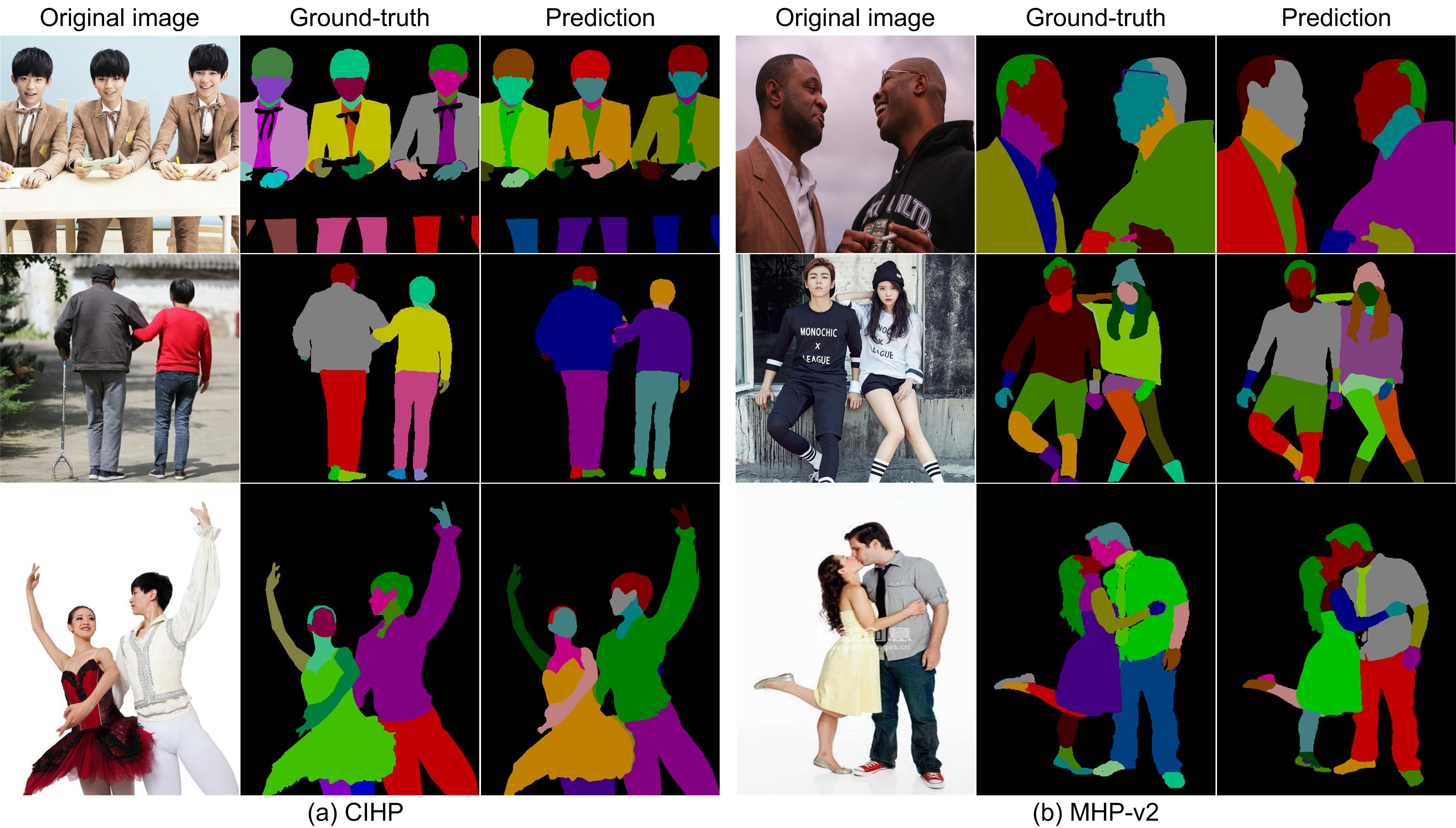}
    \caption{Some visualized examples for multiple human parsing: (a) the CIHP dataset \cite{DBLP:conf/eccv/GongLLCYL18} and (b) the MHP-v2 dataset \cite{DBLP:conf/mm/ZhaoLCSYF18}.}
    \label{fig:MHP_demo}
\end{figure*}

Furthermore, we evaluate the proposed method on two large-scale multiple human parsing datasets, \ie, CIHP \cite{DBLP:conf/eccv/GongLLCYL18} and MHP-v2 \cite{DBLP:conf/mm/ZhaoLCSYF18}. For a fair comparison, we apply same Mask R-CNN model to output instance segmentation masks. Then, we use the global parsing and two local parsing models for human parsing as in \cite{DBLP:journals/corr/abs-1809-05996}. Following the \cite{DBLP:journals/corr/abs-1809-05996}, final results are obtained by fusing the results from three branch models with a refinement process. Some visual results are shown in Figure \ref{fig:MHP_demo}, which indicates that our method can also generate precise and fine-grained results in multiple human parsing scenes.

{\noindent \textbf{Evaluation on CIHP Dataset.}}
The CIHP dataset \cite{DBLP:conf/eccv/GongLLCYL18} is the largest multi-person human parsing dataset with $38,280$ diverse human images, \ie, $28,280$ training, $5,000$ validation and $5,000$ test images. It is labeled with pixel-wise annotations on $20$ categories and instance-level identification. We use the same topology (\ie, $3$-level tree structure as shown in Figure \ref{fig:structure}) in the LIP dataset \cite{DBLP:journals/pami/LiangGSL19} to perform human parsing because the two datasets share the same sub-category semantic annotations.

As shown in Table \ref{tab:CIHP}, our method outperforms other compared methods (\ie, PGN \cite{DBLP:conf/eccv/GongLLCYL18} and M-CE2P \cite{DBLP:journals/corr/abs-1809-05996}), achieving $\text{AP}^r_m$ score of $43.96$. It is worth mentioning that SNT outperforms M-CE2P \cite{DBLP:journals/corr/abs-1809-05996} in terms of $\text{AP}^r_{0.7}$ score by considerable improvement, \ie, $29.74$ vs. $33.00$. It indicates that our method facilitates improving the segmentation accuracy of human instances.

\begin{table}[t]
\centering
		\setlength{\tabcolsep}{5.5pt}		
		\caption{The evaluation results on the validation set of CIHP \cite{DBLP:conf/eccv/GongLLCYL18}. $\text{AP}^r_m$ denotes the mean value.}
		\vspace{2mm}
		\label{tab:CIHP}
		\small{
		\begin{tabular}{ccccccccc}
			\toprule
			Method     & \text{mIoU}        & $\text{AP}^r_{0.5}$   & $\text{AP}^r_{0.6}$    & $\text{AP}^r_{0.7}$    & $\text{AP}^r_m$     \\
			\midrule
			PGN \cite{DBLP:conf/eccv/GongLLCYL18}        & 55.89       & 35.80      & 28.60          & 20.50          & 33.60     \\
			M-CE2P \cite{DBLP:journals/corr/abs-1809-05996}     & 59.50       & 48.69      & 40.13          & 29.74          & 42.83     \\
			\midrule
			Ours       & \bf{60.87}       & \bf{49.27}      & \bf{41.98}          & \bf{33.00}          & \bf{43.96}     \\
			\bottomrule
		\end{tabular}
		}
\end{table}

{\noindent \textbf{Evaluation on MHP-v2 Dataset.}}
The MHP-v2 dataset \cite{DBLP:conf/mm/ZhaoLCSYF18} includes $25,403$ elaborately annotated images with $58$ fine-grained semantic category labels, involving $2\sim26$ persons per image and captured in real-world scenes from various viewpoints, poses, occlusion, interactions and background. Since this dataset has more labels than the LIP dataset \cite{DBLP:journals/pami/LiangGSL19}, we construct the tree architecture in $5$-level.

As shown in Table \ref{tab:MHP-v2}, the semantic segmentation method Mask R-CNN \cite{DBLP:conf/iccv/HeGDG17} only obtains $\text{PCP}_{0.5}$ score of $25.12$ and $\text{AP}^p_{0.5}$ score of $14.50$ on the challenging multiple human parsing dataset. NAN \cite{DBLP:journals/corr/abs-1804-03287} achieves the best $\text{AP}^p_m$ score of $42.77$, but much inferior performance in both $\text{PCP}_{0.5}$ and $\text{AP}^p_{0.5}$ scores. Our method achieves comparable state-of-the-art performance whth M-CE2P \cite{DBLP:journals/corr/abs-1809-05996} in terms of three metrics. It indicates that the coarse to fine process in a hierarchical design can facilitate improving the accuracy.

\begin{table}[t]
	\begin{center}
		\setlength{\tabcolsep}{8.5pt}			
		\caption{The performance on the validation set of MHP-v2 \cite{DBLP:conf/mm/ZhaoLCSYF18}. $\text{AP}^p_m$ denotes the mean value.}
	\vspace{2mm}
    	\label{tab:MHP-v2}
		\small{			
		\begin{tabular}{cccccc}
			\toprule
			Method    										 & $\text{PCP}_{0.5}$      & $\text{AP}^p_{0.5}$   & $\text{AP}^p_m$  \\
			\midrule
			Mask R-CNN \cite{DBLP:conf/iccv/HeGDG17}		 & 25.12            & 14.50          & -		\\
			MH-Parser \cite{DBLP:journals/corr/LiangLSFYX17} & 26.91            & 18.05          & -		\\
			NAN \cite{DBLP:journals/corr/abs-1804-03287}     & 34.37       		& 24.87      	 & \textbf{42.77}    \\		
			M-CE2P \cite{DBLP:journals/corr/abs-1809-05996}  & \textbf{43.77}   & \textbf{34.47} & 42.70    \\
			\midrule			
			Ours											 & 43.50       		& 34.36      	 & 42.51    \\
			\bottomrule
	    \end{tabular}
	    }
	\end{center}
\end{table}

\subsection{Ablation study}
We study the influence of some important parameters and components of our SNT method, \ie, the height of the tree, the attention routing module, the semantic aggregation module and the predition module. The experiment is conducted on the LIP dataset \cite{DBLP:journals/pami/LiangGSL19}.

{\noindent \textbf{Height of the tree.}}
The height of the tree $k$ indicates the complexity of the network. To explore the optimal height, we design five variants with different heights of the tree, see Figure \ref{fig:structure}(a). If the height is equal to $0$, only the ResNet-101 backbone is used for human parsing. As presented in Table~\ref{tab:tree_height}, we can observe there is a sharp decline in mean accuracy and \text{mIoU} score. We find that our method with $3$-level achieves the best performance, \ie, $54.73\%$ \text{mIoU} score. This is attributed to two reasons. First, the model is not fine enough to predict the labels based on limited number of parameters in our model when the height is less than $3$, resulting in limited performance. Second, too deep tree (\ie, $k>3$) corresponds to many parameters. However, training on limited data may cause over-fitting of our model to decrease the accuracy slightly.

\begin{table}[t]
	\begin{center}
		\setlength{\tabcolsep}{1.5pt}		
		\caption{Effect of the height of the tree on the LIP dataset \cite{DBLP:journals/pami/LiangGSL19}.}
		\vspace{2mm}
		\label{tab:tree_height}
		\small{
		\begin{tabular}{cccc}
			\toprule
			height of the tree    & pixel acc. (\%)  	& mean acc. (\%)	 & mIoU (\%) \\
			\midrule
			0        & 84.81        & 	57.12		& 46.34\\			
			1        & 86.84        & 	64.03		& 52.15\\
			2        & 87.42 	  	&	65.58		& 53.32\\
			3        & \bf{88.05} 	&\bf{66.42}		&\bf{54.73}\\
			4        & 86.92        &	64.34		& 51.42\\
			\bottomrule
		\end{tabular}
		}
	\end{center}
\end{table}

{\noindent \textbf{Effectiveness of prediction module.}}
To analyze prediction module in the proposed network, we construct two variants of our method, \ie, ``ours w/o dconv'' and ``ours w/o pred''. As shown in Figure \ref{fig:structure}(d), the ``ours w/o dconv'' method indicates that we use traditional convolutional layers instead of deformable convolutional layers in the prediction module; while the ``ours w/o pred'' method indicates that we combine the prediction results of each leaf node for final parsing result without the prediction module.

From Table \ref{tab:transformer}, our method performs better than the ``ours w/o dconv'' method with $0.31\%$ improvement in terms of \text{mIoU}. It indicates that the deformable convolutional layers can facilitate align the semantic information in different channels of feature maps. If we do not use the prediction module to generate the final parsing result, we can observe a sharp decrease in \text{mIoU} score, \ie, $50.02$ vs. $54.73$. It is essential to achieve accurate parsing result based on the context information among every sub-categories.

{\noindent \textbf{Effectiveness of semantic aggregation module.}}
To verify the effectiveness of the semantic aggregation module, we construct the ``ours w/o skip'' method, which indicates that we do not combine the residual blocks from the backbone in attention aggregation (see Figure \ref{fig:structure}(c)). Based on the comparison between our method and the ``ours w/o skip'' method, we can conclude that the skip-connection from the backbone (see the dashed blue lines in Figure \ref{fig:structure}(a)) can bring $1.41\%$ \text{mIoU} improvement. This is because the skip-connection in our network can exploit multi-scale representation for sub-categories.

{\noindent \textbf{Effectiveness of attention routing module.}}
To study the effect of the attention routing module, the ``ours w/o mask'' indicates that we further remove the attention mask in the attention routing module from the ``ours w/o skip'' method (see Figure \ref{fig:structure}(b)). That is, we directly split the feature maps into several semantic maps for the next level. As presented in Table \ref{tab:transformer}, the ``ours w/o skip'' method achieves $2.58\%$ improvement in \text{mIoU} score compared with the ``ours w/o mask'' method. It demonstrates the attention mask can enforce the tree network focus on discriminative representation for specific sub-category semantic information.

\begin{table}[t]
	\begin{center}
		\setlength{\tabcolsep}{2.5pt}	
		\caption{Variants of the SNT method on the LIP dataset \cite{DBLP:journals/pami/LiangGSL19}.}
		\vspace{2mm}
		\label{tab:transformer}
		\small{
		\begin{tabular}{cccc}
			\toprule
			variant    & pixel acc. (\%)  	& mean acc. (\%)	 & mIoU (\%) \\
			\midrule
			Ours w/o mask        & 86.84        & 	64.03		& 52.15\\
			Ours w/o skip        & 87.42 	  	&	65.58		& 53.32\\
			Ours w/o pred        & 85.34        &   63.22       & 50.02\\			
			Ours w/o dconv        & 87.61        &   66.05       & 54.42\\						
			Ours        & \bf{88.05} 	 	&	\bf{66.42}		& \bf{54.73}\\
			\bottomrule
		\end{tabular}
		}
	\end{center}
\end{table}

\section{Conclusion}
In this paper, we propose a novel semantic tree network for human parsing. Specifically, the proposed tree architecture can encode physiological structure of human body and segment multiple semantic subregions in a hierarchical way. Extensive experiment on four challenging single and multiple human parsing datasets indicates the effectiveness of the proposed semantic tree structure. Our method can learn discriminative feature representation and exploit more context information for sub-categories effectively. For future work, we plan to optimize the tree architecture for better performance by neural architecture search techniques.

\appendix

\section{The Category Label Definition in the LIP and CIHP Datasets}
Since the LIP \cite{DBLP:journals/pami/LiangGSL19} and CIHP \cite{DBLP:conf/eccv/GongLLCYL18} datasets use the same annotations, we adopt the same architecture of our neural tree, shown in Figure \ref{fig:anno_tree}. We report the category label definition in our neural tree in Table \ref{tab:lip_label}.

\begin{table}[h]
	\caption{The category label definition used in the LIP \cite{DBLP:journals/pami/LiangGSL19} and CIHP \cite{DBLP:conf/eccv/GongLLCYL18} datasets.}
	\label{tab:lip_label}
	\begin{center}
		\begin{tabular}{|c|c|}
			\hline
			Leaf & Label \\ \hline
			$L^1_1$& Background \\ \hline
			$L^2_1$ & Hat, Hair, Sunglasses, Face \\ \hline
			$L^3_1$ & Scarf, Upper-clothes, Coat, Dress\\ \hline
			$L^3_2$ & Left-arm, Right-arm, Glove\\ \hline
			$L^3_3$ & Skirt, Pants, Jumpsuit, Left-leg, Right-leg \\ \hline
			$L^3_4$ & Socks, Left-shoe, Right-shoe\\
			\hline
		\end{tabular}
	\end{center}
\end{table}

\section{The Architecture of Our Semantic Neural Tree in the Pascal-Person-Part Dataset}
\begin{figure*}[t]
    \centering
    \includegraphics[width=0.9\linewidth]{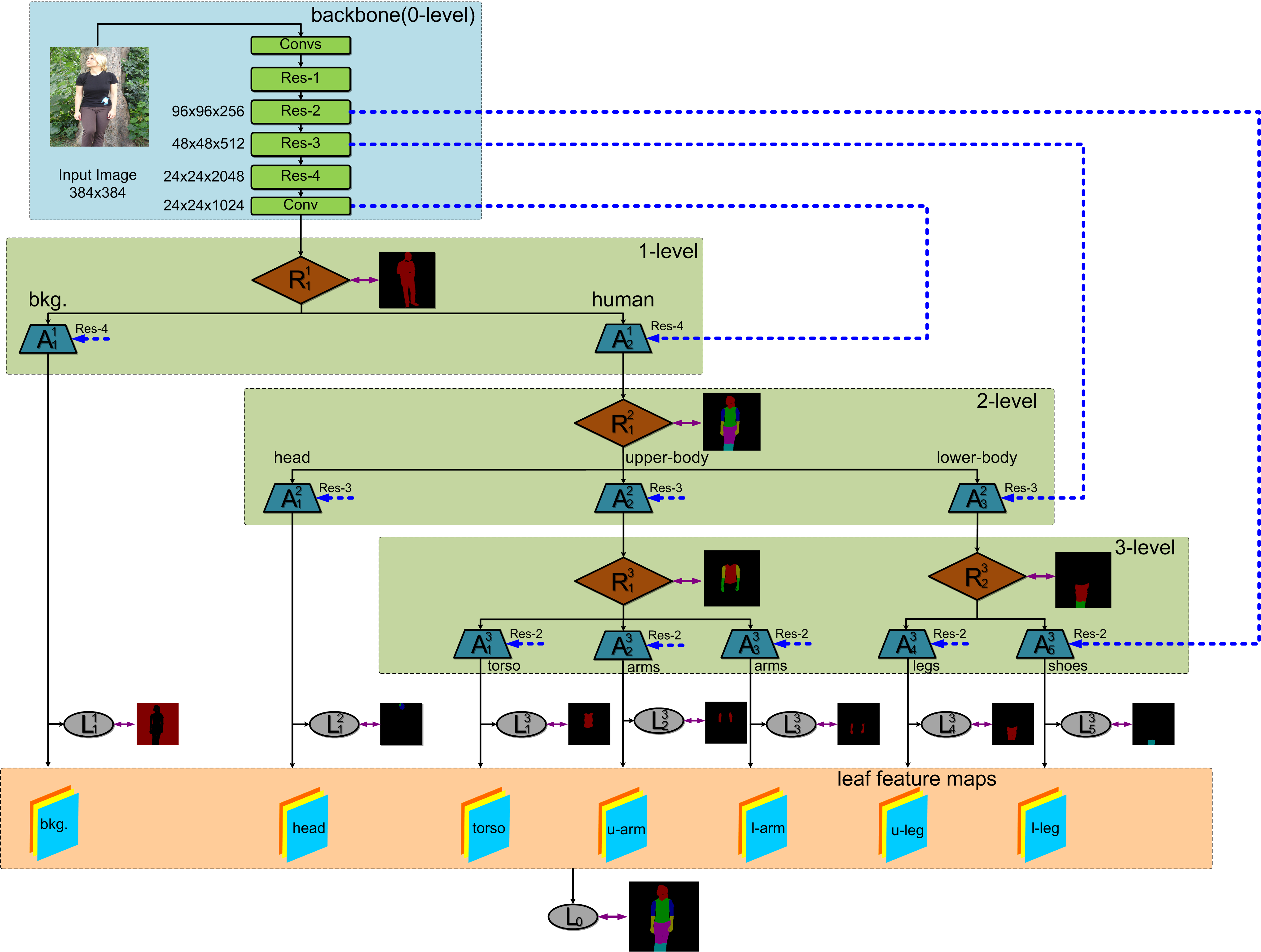}
    \caption{The architecture of our semantic neural tree used in the Pascal-Person-Part dataset \cite{DBLP:conf/cvpr/ChenMLFUY14}.}
    \label{fig:ppp_structure}
\end{figure*}

The PASCAL-Person-Part dataset \cite{DBLP:conf/cvpr/ChenMLFUY14} is originally from the PASCAL VOC-2010 dataset \cite{DBLP:journals/ijcv/EveringhamGWWZ10}, and is extended for human parsing with $6$ coarse body part labels (\ie, \textit{head}, \textit{torso}, \textit{upper-/lower-arms}, and \textit{upper-/lower-legs}). As shown in Figure \ref{fig:ppp_structure}, we construct a neural tree with $3$-level. We summarize the category label definition in Table \ref{tab:ppp_label}.

\begin{table}[h]
	\caption{The category label definition used in the Pascal-Person-Part dataset.}
	\label{tab:ppp_label}
	\begin{center}
		\begin{tabular}{|c|c|}
			\hline
			Leaf & Label \\ \hline
			$L^1_1$& Background \\ \hline
			$L^2_1$ & Head   \\ \hline
			$L^3_1$ & Torso  \\ \hline
			$L^3_2$ & U-arms \\ \hline
			$L^3_3$ & L-arms \\ \hline
			$L^3_4$ & U-legs \\ \hline
			$L^3_5$ & L-legs \\
			\hline
		\end{tabular}
	\end{center}
\end{table}

\section{The Architecture of Our Semantic Neural Tree in the MHP-v2 Dataset}
\begin{figure*}[t]
    \centering
    \includegraphics[width=0.9\linewidth]{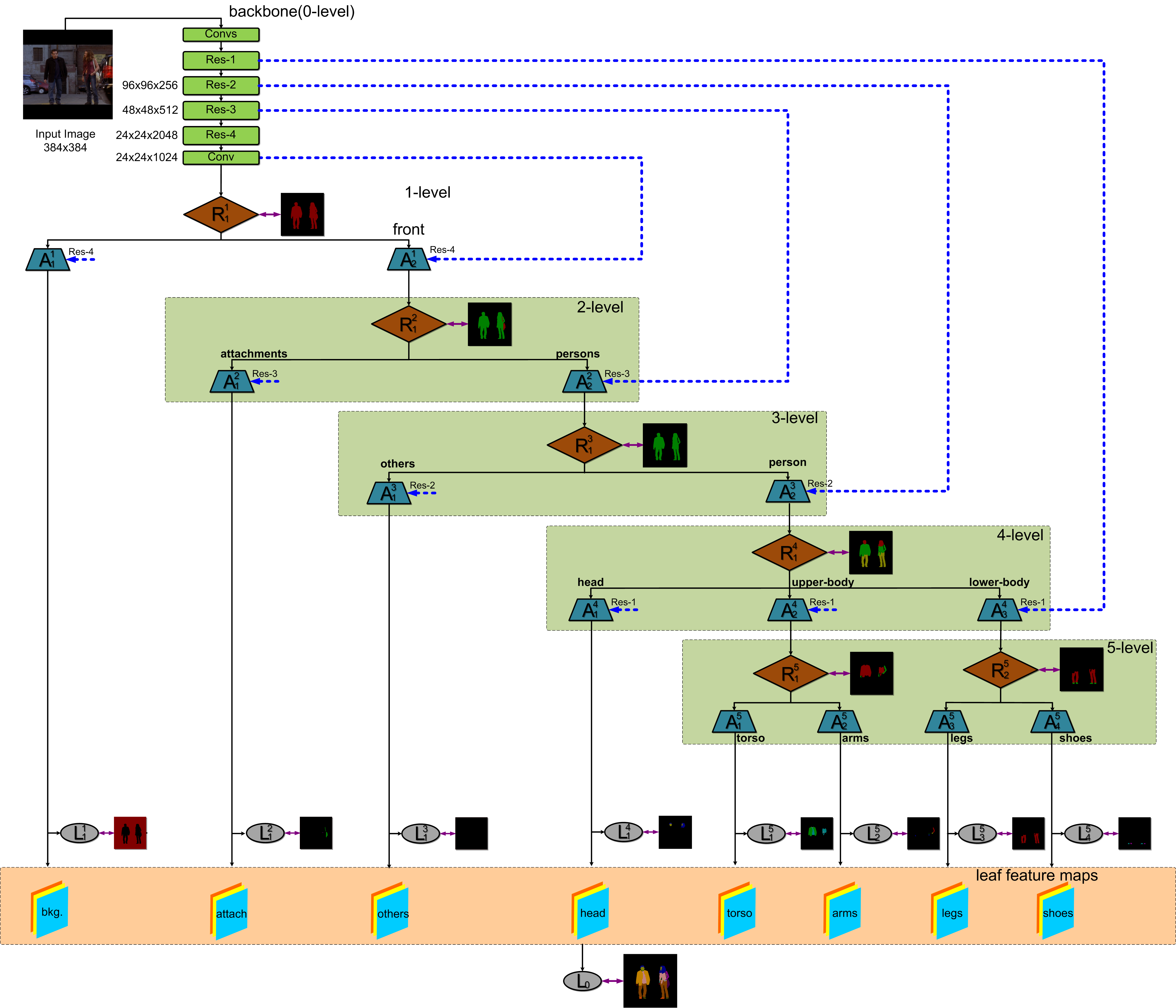}
    \caption{he architecture of our semantic neural tree used in the MHP-v2 dataset \cite{DBLP:conf/mm/ZhaoLCSYF18}.}
    \label{fig:mhp_structure}
\end{figure*}

The MHP-v2 dataset \cite{DBLP:conf/mm/ZhaoLCSYF18} includes $25,403$ elaborately annotated images with $58$ fine-grained semantic category labels, involving $2\sim26$ persons per image and captured in real-world scenes from various viewpoints, poses, occlusion, interactions and background. As shown in Figure \ref{fig:mhp_structure}, we construct a neural tree with $5$-level. We summarize the category label definition in Table \ref{tab:mhp_label}.

\begin{table}[h]
	\setlength{\tabcolsep}{0.5pt}
	\caption{The category label definition used in the MHP-v2 dataset.}
	\label{tab:mhp_label}	
	\footnotesize
	\begin{center}
		\begin{tabular}{|c|c|}
			\hline
			Leaf & Label \\ \hline
			$L^1_1$& Background \\ \hline
			$L^2_1$ & Backpack, Protector, Ball, Bats, Bottle, Carrybag, Cases,\\ & Umbrella, Wallet/Purse   \\ \hline
			$L^3_1$ & Other-full-body-clothes, Other-accessary, Other-upper-body-clothes, \\& Other-lower-body-clothes  \\ \hline
			$L^4_1$ & Cap/Hat, Helmet, Hair, Sunglasses, Face, Headwear, Eyewear \\ \hline
			$L^5_1$ & Bikini/Bra, Jacket/Windbreaker/Hoodie, Tee-shirt, Polo-shirt, \\& Sweater, Singlet, Torso-skin, Robe, Coat, Dress, Tie, Scarf, Belt \\ \hline
			$L^5_2$ & Glove, Watch, Wristband, Left-arm, Right-arm, Left-hand, Right-hand\\ \hline
			$L^5_3$ & Left-leg, Right-leg, Jumpsuit, Pants, Shorts/Swim-shorts, Skirt \\ \hline
			$L^5_4$ & Stockings, Socks, Left-boot, Right-boot, Left-shoe, Right-shoe \\&Left-higheel, Right-higheel, Left-sandal, Right-sandal, Left-foot,\\& Right-foot \\
			\hline
		\end{tabular}
	\end{center}
\end{table}

{\small
\bibliographystyle{ieee_fullname}
\bibliography{references}
}

\end{document}